\documentclass[11pt]{article}

\usepackage[]{acl}

\usepackage{times}
\usepackage{latexsym}

\usepackage[T1]{fontenc}

\usepackage[utf8]{inputenc}

\usepackage{microtype}

\usepackage{xcolor}

\usepackage{array}
\usepackage{amssymb}
\usepackage{booktabs}
\usepackage{colortbl}
\usepackage{comment}
\usepackage{graphicx}
\usepackage{tikz-dependency}
\usepackage{makecell}
\usepackage{multirow}
\usepackage{amsmath}

\usepackage{blindtext}
\usepackage{varwidth}
\usepackage{tabularx}

\graphicspath{ {./charts/} }

\DeclareMathAlphabet\mathbfcal{OMS}{cmsy}{b}{n}

\title{The Fragility of Multi-Treebank Parsing Evaluation}

  \author{
  Iago Alonso-Alonso, David Vilares and Carlos G\'{o}mez-Rodr\'{i}guez \\
  Universidade da Coru\~{n}a, CITIC \\
  Departamento de Ciencias de la Computación y Tecnologías de la Información \\
  Campus de Elvi\~{n}a s/n, 15071 \\ A Coru\~{n}a, Spain \\
  {\tt \{iago.alonso,david.vilares,carlos.gomez\}@udc.es} 
  \\}

\date{}

\begin{document}
\maketitle
\begin{abstract}

Treebank selection for parsing evaluation and the spurious effects that might arise from a biased choice have not been explored in detail. This paper studies how evaluating on a single subset of treebanks can lead to weak conclusions. First, we take a few contrasting parsers, and run them on subsets of treebanks proposed in previous work, whose use was justified (or not) on criteria such as typology or data scarcity. Second, we run a large-scale version of this experiment, create vast amounts of random subsets of treebanks, and compare on them many parsers whose scores are available. The results show substantial variability across subsets and that although establishing guidelines for good treebank selection is hard, it is possible to detect potentially harmful strategies.
\end{abstract}

\section{Introduction}

A limitation in NLP evaluation lies in the association between solving a dataset \emph{versus} solving a task. Datasets are domain-specific, their sizes differ and they are only available for a handful of languages and cultures \cite{hershcovich-etal-2022-challenges}. 
Yet, we often ignore that the chances that these results generalize in the real world are scarce. In this context, the conclusions extracted from a single dataset should be taken with caution.

For dependency parsing, the Universal Dependencies framework \citep[UD;][]{11234/1-3424} mitigates some of these issues. For instance, version 2.8 of UD includes 202 treebanks and 114 languages covering diverse linguistic typologies, treebanks with different amounts of data, and domains. Paradoxically, this 
also complicates decisions when it comes to comparing dependency parsers in multilingual environments, which can be summarized as: \emph{how to choose a small but representative set of treebanks?} Although there are shared tasks \cite{zeman-etal-2017-conll,zeman-etal-2018-conll} that do consider experiments over a wide set of treebanks and help understand parsing models, such setups do not usually stick when the shared tasks end, and authors often run their models only in a handful of treebanks \citep[\emph{inter alia}]{deLhoneux2016UDTS,ma-etal-2018-stack,kulmizev-etal-2019-deep}. 
This mostly happens for justified reasons: lack of computational resources to train the models in a reasonable amount of time, energy usage concerns, difficulties to summarize large experiments, or interest in specific phenomena (e.g. non-projectivity). Thus, a good treebank selection strategy is crucial to reduce the chances of selecting an unrepresentative subset of treebanks, which could lead to weak conclusions. 
Furthermore, even when using the whole UD collection is viable, treebank selection can still be relevant as UD is not a representative sample of languages (e.g., 62 out of the 114 languages in v2.8 are Indo-European), so coarse-grained measures like averages over all treebanks may be misleading. 

\paragraph{Contribution} We hypothesize that using a single subset of treebanks can be a weak approach to extract conclusions about 
the performance of parsers and their rankings.
To test so, we design two experiments.
First, we choose representative models of different paradigms: a graph-based \cite{dozat-etal-2017-stanfords}, a transition-based \cite{fernandez-gonzalez-gomez-rodriguez-2019-left}, and a sequence tagging \cite{strzyz-etal-2019-viable} parser; and evaluate them on a few subsets defined in the literature, looking for different trends.
Then, we redefine the previous experiment on a large scale. We take the output of dozens of parsers on the treebanks used at the UD CoNLL 2018 shared task \cite{zeman-etal-2018-conll} to study the variability of parsing rankings over a million of fixed-size, randomly generated subsets.

\section{Related work}

The appropriateness of experimental setups for parsing evaluation 
has been studied in recent years from different perspectives.

Some authors have focused on determining what are the treebank particularities that make some of them easier to parse than others. For instance, the size of the training set is widely known to be an important factor to obtain accurate results in dependency parsing \cite{dehouck-denis-2019-phylogenic,vania-etal-2019-systematic}. 
Other aspects such as domain similarity \cite{wisniewski-yvon-2019-bad} or annotation similarity \cite{dredze-etal-2007-frustratingly, cohen-etal-2012-domain}  between the training and test sets have also been studied, showing that they can greatly affect the performance of parsers. Other particularities that can also affect the performance on a treebank are linguistic variation \cite{nivre-etal-2007-conll}, 
annotation criteria \cite{kubler-etal-2008-compare,rosa-2015-multi}, arc direction \cite{rehbein-etal-2017-universal}, average dependency length \cite{gulordava-merlo-2016-multi}, non-projectivity \cite{kuhlmann2010transition}, morphological richness \cite{tsarfaty-etal-2013-parsing} or information-theoretic metrics \cite{Corazza2013}, among other factors. 

Although not specifically for parsing but NLP, \citet{gorman-bedrick-2019-need} and \citet{sogaard-etal-2021-need} comment that the way data is split can play a role on 
test results, and thus on conclusions. Extrapolating this to parsing, it would suggest that some parsers could obtain better results for certain treebanks just 
due to data splitting decisions, and not due to a linguistic motivation that would explain a given language being harder to parse than other.  
Recently, \citet{sogaard-2020-languages} 
studied the influence of overlap between trees in training and test sets in a given split, and
concluded that (the amount of) graph isomorphism between the training and test set trees partially explains why some treebanks are easier or harder to parse than others. However, \citet{anderson-etal-2021-replicating} replicated the study, controlling for covariants, and 
proved that much of this observation is explained by relevant covariants like treebank size and mean test sentence length.

Another line of research more related to our work involves the studies that compare how different parsing algorithms behave on the same held-out test sets. \citet{mcdonald-nivre-2007-characterizing,mcdonald-nivre-2011-analyzing} showed that non-neural transition-based and graph-based parsers perform overall similarly, but produce different types of errors, with transition-based parsers being weaker for long dependencies and graph-based parsers weaker for shorter, more local ones. Relatedly, \citet{de2017old} compared a neural and non-neural transition based parser, showing that the former is not only clearly better at longer dependencies, but that it also needs less training data to parse effectively. 
\citet{kulmizev-etal-2019-deep} replicated the work by \citeauthor{mcdonald-nivre-2011-analyzing} for neural versions of those parsers and, contrarily, demonstrated that the contextualization of the input vectors with recurrent networks results into both types of parsers showing a much more homogeneous behavior. Also related to this, \citet{anderson-gomez-rodriguez-2020-inherent} showed how different transition-based algorithms are prone to outperform others on a specific treebank according to their inherent dependency displacement biases. 

To the best of our knowledge, there have been only two papers in the literature that specifically focus on presenting methodologies to choose a suitable set of treebanks for parsing evaluation, both centered on UD and with the goal of obtaining a small sample of treebanks that is representative of the full UD collection (not necessarily of human languages as a whole). \citet{deLhoneux2016UDTS,Lhoneux2017OldSV} do so by manually selecting treebanks to enforce typological diversity as well as representativity in other relevant aspects for parsing, like projectivity or treebank size. In turn, \citet{schluter-agic-2017-empirically} take an automatic, quantitative approach, obtaining a sample by clustering using delexicalized parsing performance. While many other papers have presented and used 
subsets of UD treebanks for evaluation, they either do not focus on representativity (e.g. \citet{ma-etal-2018-stack}) or follow one of these methodologies (e.g. \citet{anderson-gomez-rodriguez-2020-distilling}).

\section{Hyphothesis and methodology}

As suggested above, parsing conclusions on multilingual environments are usually drawn from empirical research, which are prone to be parser-specific, experiment-specific, as well as treebank dependent. 

\paragraph{Hypothesis} 
We delve into this problem and hypothesize that parser comparisons based on running experiments and taking accuracy metrics on a given (reasonably-sized) subset of treebanks may lead to weak conclusions on rankings or differences in performance; as the magnitude and/or sign of the differences between parsers can change substantially depending on the choice of said subset.

\subsection{Methodology} To test our hypothesis, we design two experiments:

\paragraph{Experiment 1: \emph{few} controlled parsers, \emph{few} pre-existing subsets} In \S \ref{section-experiment-1}, we choose three contrasting parsers belonging to different parsing paradigms. Then, we train and evaluate them on a number of pre-defined (multilingual) subsets that were proposed in previous work (and later adopted by other authors as well). These existing subsets present different particularities, such as a high Indo-European bias \cite{ma-etal-2018-stack}, rich and diverse typologies \cite{Lhoneux2017OldSV}, or data scarcity issues \cite{anderson-et-al-2021}, among others. Our aim is to see whether considering only a few robust parsers (treated as black boxes) and only a few already established subsets of treebanks, we can obtain different conclusions about their behaviors.
    
\paragraph{Experiment 2: \emph{many} parsers, \emph{many} randomized subsets} In \S \ref{section-experiment-2}, we design a large-scale variant of the previous experiment. 
Assuming access to many parsers and treebanks, we ask: \emph{could we obtain (reasonably-sized) subsets of treebanks that show very different behaviors?}, or to state it differently, \emph{can  parsing rankings be sensitive to the subset of treebanks where they are evaluated?} To do so, we use as a proxy the results from the CoNLLU Shared Task 2018 \cite{zeman-etal-2018-conll}, where 26 parsers participated and presented their experiments for 82 treebanks. We then create a random sample of 1 million subsets out of the $\sim 2.13\times 10^{12}$ possible (multilingual) subsets of size 10. If a parser is cross-linguistically robust, then the variability of its position in the ranking should be small across all the studied subsets, while if their behavior is more unstable it could change dramatically, 
indicating that evaluating on a single subset of treebanks is not desirable.

\section{Experiment 1: \emph{few} controlled parsers, \emph{few} pre-existing subsets}\label{section-experiment-1}

We take a few representative parsers (\S \ref{section-parsing-models}) and
pre-defined subsets 
from the literature 
(\S \ref{section-datasets-exp1}) based, sometimes, on a careful treebank selection strategy. 

\subsection{The parsing models}\label{section-parsing-models}

We choose a graph-based \cite{dozat-etal-2017-stanfords}, a transition-based \cite{fernandez-gonzalez-gomez-rodriguez-2019-left}, and a sequence labeling parser \cite{strzyz-etal-2019-viable}. We review them briefly, but we refer the reader to the papers for the details.

\paragraph{Bi-affine graph-based parser} \citep[\texttt{gb-DM17};][]{dozat-etal-2017-stanfords} 
It first computes contextualized vectors for each word using bidirectional LSTMs \citep[biLSTMs;][]{hochreiter1997long}. After that, the model computes for each word a \emph{head} and a \emph{dependent} representation, which are sent through a bi-affine attention, determining for each token which is the most likely head. 
Here, we rely on the \texttt{supar}\footnote{\url{https://github.com/yzhangcs/parser}} package, which has been widely adopted by the community. We detail its hyperparameters in Appendix \ref{appendix-hyperparameters} (Table \ref{tab:hyperparameters-supar}).

\paragraph{Left-to-right, transition-based, pointer network parser} \citep[\texttt{tb-FG19};][]{fernandez-gonzalez-gomez-rodriguez-2019-left} 
It is a transition-based system, where at each time-step the pointer network predicts the index of the head for the focus token, and moves to the next one.
The model uses an encoder-decoder architecture that in the first stage computes a hidden state representation for each token using biLSTMs. After that, the decoder predicts the tree left to right, computing a score attention between the current focus word  and the encoder output sequence, excluding the own vector. We use the \texttt{syntacticpointer}\footnote{\url{https://github.com/danifg/SyntacticPointer}} package. 
Appendix \ref{appendix-hyperparameters} (Table \ref{tab:hyperparameters-syntacticPointer}) details the hyperparameters.

\paragraph{Sequence labeling parser} \citep[\texttt{sl-S+19};][]{strzyz-etal-2019-viable}
It outputs a dependency tree for each sentence of size $n$, using exactly $n$ predictions and biLSTM tagging models. There are different ways to encode the trees \cite{spoustova2010dependency,lacroix-2019-dependency,gomez-rodriguez-etal-2020-unifying}, but we will rely on the 2-planar bracketing encoding \cite{strzyz-etal-2020-bracketing}, which encodes 99.9\% of non-projective trees and offers a robust behavior, including low-resource setups \cite{munoz-ortiz-etal-2021-linearizations}.
We use the \texttt{dep2labels}\footnote{\url{https://github.com/mstrise/dep2label}} package. The hyperparameters are indicated in Appendix \ref{appendix-hyperparameters} (Table \ref{tab:hyperparameters-dep2label}).

\paragraph{Parser comparability} 

Sequence labeling parsers often underperform biaffine and pointer network parsers \cite{anderson-gomez-rodriguez-2021-modest}, but we include them as a lower bound control parser.
We kept fundamental architectural decisions of the parsers, e.g. how they compute character-level vectors or the strategies for cycle deletion, as it is not clear (or viable) that the same setup is optimal for all models. Also, we value to try these models as used by the community.

\subsubsection{Experiment setup}

\noindent The parsers are trained 3 times for each treebank, to then take the average as the final result.\footnote{Some treebanks (Kazakh\textsubscript{KTB}, Galician\textsubscript{TreeGal} and Old-East-Slavic\textsubscript{RNC}) do not have an official dev set, so we used 20\% of the training set as the development set.
Also, due to hardware limitations, the longest sentence (682 tokens) in the test file for the Old East Slavic (RNC) language was removed, since \texttt{syntacticpointer} ran out of memory during evaluation.}\textsuperscript{,}\footnote{The tools used for the experiments are described at this repository: \url{https://github.com/MinionAttack/fragility_coling_2022}}

\paragraph{Input} For all parsers, the embedding for each word is composed of a pre-trained word vector, a character-based vector and a PoS tag vector. 
For the word vectors, we use \texttt{fastText} \cite{bojanowski2017enriching}.\footnote{We use \texttt{fastText} vectors except for some language treebanks that lacked embeddings. Particularly, for Ancient Greek we use UD embeddings, and for Wolof we used random initialized embeddings according to a uniform distribution in the range $[\frac{-1}{2 \times 300}, \frac{1}{2 \times 300}]$ \cite{goldberg2017neural}.} For PoS tags, we considered experiments both with gold and predicted PoS tags - using UDpipe \cite{straka-etal-2016-udpipe}\footnote{For Kazakh, Old East Slavic and Welsh there are no UDPipe models, so we only include their results with gold tags.}.

\subsection{Datasets}\label{section-datasets-exp1}

Now, we review subsets that have been proposed, and summarize the criteria used to create them.
While the subsets were defined on different versions of UD depending on the moment in which they were proposed, we use
UD v2.8 for comparability. In case different treebanks are available for a given language and the authors did not specify which one they used for any reason (e.g. because in previous UD versions there was only one treebank, and therefore it was not necessary to name it), we chose the largest freely-available one. For space reasons, we include the specific treebanks for each subset in Appendix \ref{appendix-treebanks-per-subset} (Table \ref{tab:treebanks-per-set}).

\begin{table*}[thbp]
\centering
\addtolength{\tabcolsep}{-2pt}
\scriptsize
 \begin{tabular}{l|ccc|ccc|ccc|ccc}
  \multirow{3}{*}{\textbf{Set}} 
  & \multicolumn{1}{c}{} & \multicolumn{1}{c}{\textbf{LAS}} & \multicolumn{1}{c|}{} & \multicolumn{3}{c|}{$\mathbfcal{E}$-LAS} & \multicolumn{1}{c}{} & \multicolumn{1}{c}{\textbf{UAS}} & \multicolumn{1}{c|}{} & \multicolumn{1}{c}{} & \multicolumn{1}{c}{\textbf{$\mathbfcal{E}$-UAS}} & \multicolumn{1}{c}{} \\ 
  \multicolumn{1}{c|}{} &
  \multicolumn{1}{c @{}}{\texttt{gb-DM17}} &
  \multicolumn{1}{c @{}}{\texttt{tb-FG19}} &
  \multicolumn{1}{c @{} |}{\texttt{sl-S+19}} &
  \multicolumn{1}{>{\centering\arraybackslash}m{1.0cm}}{(\texttt{tb-FG19}, \texttt{gb-DM17})} &
  \multicolumn{1}{>{\centering\arraybackslash}m{1.0cm}}{(\texttt{sl-S+19}, \texttt{gb-DM17})} &
  \multicolumn{1}{>{\centering\arraybackslash}m{1.0cm}|}{(\texttt{sl-S+19}, \texttt{tb-FG19})} &
  \multicolumn{1}{c @{}}{\texttt{gb-DM17}} &
  \multicolumn{1}{c @{}}{\texttt{tb-FG19}} &
  \multicolumn{1}{c @{} |}{\texttt{sl-S+19}} &
  \multicolumn{1}{>{\centering\arraybackslash}m{1.0cm}}{(\texttt{tb-FG19}, \texttt{gb-DM17})} &
  \multicolumn{1}{>{\centering\arraybackslash}m{1.0cm}}{(\texttt{sl-S+19}, \texttt{gb-DM17})} &
  \multicolumn{1}{>{\centering\arraybackslash}m{1.0cm}}{(\texttt{sl-S+19}, \texttt{tb-FG19})} \\
 \hline
 Ma18 
 & 87.74 & 87.77 & 83.96 & -0.14 & 23.93 & 23.93 & 91.07 & 91.13 & 87.68 & -0.33 & 27.98 & 28.16 \\ 
 Lh16 
 & 80.33 & 79.68 & 74.20 & 1.83 & 24.04 & 22.49 & 85.03 & 84.45 & 79.90 & 2.21 & 26.33 & 24.51 \\ 
 SA17 
 & 84.85 & 84.97 & 80.30 & -1.03 & 22.97 & 23.48 & 89.11 & 89.25 & 85.22 & -1.46 & 26.76 & 27.64 \\
 Sm18 
 & 83.78 & 83.61 & 78.55 & 1.11 & 24.21 & 23.35 & 87.38 & 87.31 & 83.19 & 0.45 & 25.12 & 24.82 \\ 
 Ku19 
 & 83.36 & 83.08 & 77.98 & -0.29 & 24.79 & 24.50 & 87.43 & 87.03 & 83.19 & 0.94 & 24.89 & 23.72 \\ 
 AG20
 & 76.14 & 75.26 & 69.36 & 3.01 & 23.01 & 20.48 & 82.49 & 81.69 & 76.83 & 3.83 & 25.04 & 21.95 \\ 
 D21 
 & 59.00 & 57.04 & 51.38 & 4.57 & 17.00 & 12.97 & 68.60 & 67.38 & 62.96 & 3.94 & 16.18 & 12.92 \\ 
 Easy 
 & 89.59 & 89.65 & 85.88 & -0.63 & 25.90 & 26.42 & 92.42 & 92.55 & 89.33 & -1.58 & 28.61 & 29.85 \\  
\end{tabular}
\caption{Average LAS and UAS scores for each subset in the predicted PoS tags setup. $\mathcal{E}(M1,M2)$ stands for error reduction between two models, where M1 is the reference system.}
\label{tab:las-uas-subset-means-udpipe}
\end{table*}

\begin{table*}[thbp!]
\centering
\addtolength{\tabcolsep}{-2pt}
\scriptsize
 \begin{tabular}{l|c @{\hspace{0.5\tabcolsep}} c @{\hspace{0.5\tabcolsep}} c|ccc|ccc|ccc}
  \multirow{3}{*}{\textbf{Set}} 
  & & \textbf{LAS} & & & \textbf{$\mathbfcal{E}$-LAS} & & & \textbf{UAS} & & & \textbf{$\mathbfcal{E}$-UAS} & \\ 
  &
  \texttt{gb-DM17}  &
  \texttt{tb-FG19}  &
  \texttt{sl-S+19}  &
  \multicolumn{1}{>{\centering\arraybackslash}m{1.0cm}}{(\texttt{tb-FG19}, \texttt{gb-DM17})} &
  \multicolumn{1}{>{\centering\arraybackslash}m{1.0cm}}{(\texttt{sl-S+19}, \texttt{gb-DM17})} &
  \multicolumn{1}{>{\centering\arraybackslash}m{1.0cm}|}{(\texttt{sl-S+19}, \texttt{tb-FG19})} &
  \multicolumn{1}{c @{}}{\texttt{gb-DM17}} &
  \multicolumn{1}{c @{}}{\texttt{tb-FG19}} &
  \multicolumn{1}{c @{} |}{\texttt{sl-S+19}} &
  \multicolumn{1}{>{\centering\arraybackslash}m{1.0cm}}{(\texttt{tb-FG19}, \texttt{gb-DM17})} &
  \multicolumn{1}{>{\centering\arraybackslash}m{1.0cm}}{(\texttt{sl-S+19}, \texttt{gb-DM17})} &
  \multicolumn{1}{>{\centering\arraybackslash}m{1.0cm}}{(\texttt{sl-S+19}, \texttt{tb-FG19})} \\
 \hline
 Ma18 
 & 90.51 & 90.06 & 88.29 & 4.62 & 19.29 & 15.45 & 93.10 & 92.77 & 91.00 & 4.70 & 23.59 & 19.89 \\ 
 Lh16 
 & 78.89 & 76.71 & 74.44 & 7.20 & 19.94 & 13.55 & 84.30 & 83.13 & 80.84 & 6.10 & 21.24 & 16.13 \\ 
 SA17 
 & 85.52 & 84.57 & 81.92 & 5.71 & 20.79 & 15.89 & 89.52 & 88.83 & 86.65 & 5.32 & 23.40 & 19.00 \\
 Sm18 
 & 87.42 & 86.74 & 83.01 & 4.89 & 25.63 & 21.86 & 89.89 & 89.36 & 86.07 & 4.82 & 26.93 & 23.32 \\ 
 Ku19 
 & 87.18 & 86.14 & 82.99 & 6.22 & 24.65 & 19.58 & 89.82 & 89.04 & 86.18 & 5.87 & 26.13 & 21.52 \\ 
 AG20 
 & 81.04 & 79.54 & 77.53 & 7.17 & 14.08 & 7.35 & 85.77 & 84.82 & 82.72 & 6.46 & 16.26 & 10.50 \\ 
 D21 
 & 67.99 & 63.74 & 67.30 & 11.71 & 4.14 & -8.82 & 75.26 & 72.61 & 75.52 & 9.92 & 2.14 & -8.73 \\ 
 Easy 
 & 92.62 & 92.10 & 89.50 & 6.71 & 29.81 & 24.84 & 94.75 & 94.41 & 92.19 & 6.13 & 32.76 & 28.41 \\  
\end{tabular}
\caption{Average LAS and UAS scores for each subset in the gold PoS tags setup. 
}
\label{tab:las-uas-subset-means-ud}
\end{table*}

\begin{enumerate}

\item \citet{ma-etal-2018-stack} subset (Ma18): It has been widely adopted \citep[\emph{inter alia}]{fernandez-gonzalez-gomez-rodriguez-2019-left,li2020global,yang2021headed}, but it presents two weaknesses: (i) a high presence of Indo-European treebanks, ignoring diverse typologies, and (ii) as reported in their paper, all these treebanks are \emph{easy}\footnote{We use \emph{easy} in an informal sense, referring to treebanks where parsers obtain a higher performance. In \emph{no way} we relate this term with a language being easier than other.} treebanks.

\item \citet{deLhoneux2016UDTS,Lhoneux2017OldSV} subset (Lh16): They were the first to address the problem of selecting a diverse sample of UD treebanks, establishing the following requirements: (i) include only one treebank from coarse-grained language families, (ii) include treebanks with certain morphological particularities, (iii) ensure different amounts of data, and (iv) include at least a highly non-projective treebank.

\item \citet{schluter-agic-2017-empirically} subset (SA17): Rather than manually choosing treebanks, this subset was chosen by an empirical method based on using delexicalized parsing performance to construct a similarity network, cluster it, and take one representative of each cluster. They concluded that their subset overestimates performance, while that of \citet{deLhoneux2016UDTS} underestimates it.

\item \citet{smith-etal-2018-investigation} subset (Sm18): The selection criteria for this subset were inspired in the criteria of \citet{deLhoneux2016UDTS}, but in this case aiming to be representative of different writing systems, character set sizes, and morphological complexity.

\item \citet{kulmizev-etal-2019-deep} subset (Ku19): The authors selected 13 treebanks, inspired in the criteria by \citet{deLhoneux2016UDTS} and \citet{smith-etal-2018-investigation}. Apart from script, character set size and morphological complexity, they also aimed to have a representation of different training sizes and domains, and selected treebanks with good annotation quality.

\item \citet{anderson-gomez-rodriguez-2020-distilling} subset (AG20): Highly inspired by \citet{Lhoneux2017OldSV}, but with
a few changes.
First, they exchanged Kazakh\textsubscript{KTB} for Uyghur\textsubscript{UDT}, as Kazakh lacked an official development set. Second, they exchanged Ancient Greek\textsubscript{PROIEL} for Ancient Greek\textsubscript{Perseus}, since it's more non-projective. Third, Czech\textsubscript{PDT} is swapped with Russian\textsubscript{GSD}, as the Czech treebank took too long to train.
Finally, they included Wolof\textsubscript{WTB} since African languages were not present. We included it to see if partial and justified changes over a diverse treebank subset could still lead to non-negligible changes.

\item \citet{anderson-et-al-2021} subset (D21): This subset is dedicated to \emph{true} data scarce treebanks. In the case of treebanks without a dev file, the training file was split in two, with a ratio of 80-20 for the training file and the dev file.

\item Easy subset: We propose an explicit \emph{easy} subset to compare against other easy ones (e.g.~Ma18). We used the results from the CoNLL 2018 Shared Task\footnote{\url{https://universaldependencies.org/conll18/results-las.html}}, and chose the 10 treebanks with the best LAS (no repeated languages). We list them in Appendix \ref{appendix-treebanks-per-subset}.

\end{enumerate}

\subsection{Results}

Table \ref{tab:las-uas-subset-means-udpipe} shows the macro-average LAS (Labeled Attachment Score) and UAS (Unlabeled Attachment Score) results, using predicted PoS tags, for each subset, i.e. \emph{the subset}, and not the treebank, is considered as \emph{the atomic unit} for evaluation. For informative purposes, Table \ref{tab:las-uas-subset-means-ud} shows the equivalent evaluation with gold PoS tags, but we will focus on the results with predicted PoS tags, unless stated otherwise.
We also show error reduction ratios on LAS and UAS between parsers. This metric provides a better picture of differences between parsers than absolute LAS/UAS differences would, as it is less affected by treebank difficulty differences (e.g., it is much harder to achieve a given absolute LAS and UAS difference on easy treebanks than on more difficult ones, due to less available room for improvement and diminishing returns). 
The error reduction shown for each subset is calculated by first computing the error reduction for each treebank in the subset, and then averaging these error reductions (rather than by averaging the LAS/UAS for each treebank in the subset, and computing a single error reduction on that average). While this choice can cause some superficially counterintuitive phenomena like a parser having more average LAS than another but negative LAS error reduction (this happens with \texttt{tb-FG19} and \texttt{gb-DM17} on Ku19 on Table~\ref{tab:las-uas-subset-means-udpipe}), it provides the desired semantics: for example, if a parser improves LAS from 98\% to 99\% in one treebank and from 50\% to 90\% in another, on average it is removing 65\% of errors (50\% of the errors in the first corpus, 80\% in the second) and not 78.8\% which we would obtain if we computed error reduction on average LAS.

Next, we discuss factors that seem to play a role in the subset performance. 
\paragraph{Influence of parsing difficulty} From the results, easier subsets tend to correspond to larger error reductions when comparing the (state-of-the-art) parsers \texttt{gb-DM17} and \texttt{tb-FG19} with respect to \texttt{sl-S+19} (the control parser). This is most evident for the Easy subset: all parsers obtain their best performance across all subsets, and the error reductions with respect to the control parser are also the largest, for all setups. The opposite happens with the D21 subset, the hardest one. 
In this context, when optimizing for other dimensions than performance, such as speed, training efficiency or architectural simplicity, relying (exclusively) on easy treebanks could thus be a sub-optimal strategy. The sense of the decrease in performance could be larger on these easy datasets than when evaluating on random treebanks, or on more difficult cases as suggested by the results on the subsets of Lh16, 
D21,
or AG20 
to a lesser extent. On the contrary, dimensions such as the ones mentioned above are often not expected to benefit more from the particularities of easy treebanks.
Also, there are trends related to parsing difficulty between the state-of-the-art parsers \texttt{gb-DM17} and \texttt{tb-FG19}:
\texttt{gb-DM17} seems to be superior
to \texttt{tb-FG19} when the subset becomes harder to parse, and \emph{vice versa}.

\paragraph{Differences on representative subsets} 
While both the Lh16 and the SA17 subsets were designed to enforce representativity, the ranking of the
tested parsers changes: \texttt{tb-FG19} performs better (in LAS error reduction terms) than \texttt{gb-DM17} on SA17 (automatically picked) and Ku19 (manually constructed), but
worse on the (manually constructed) subsets of Lh16, Sm18 and AG20.
This highlights that even when treebanks are sampled with attention to representativity, results can still show instability - be it due to different possible notions of representativity, or statistical variation. 

\paragraph{Developing and testing on the same treebanks} 
While there is no clear performance difference between \texttt{gb-DM17} and \texttt{tb-FG19}, as each of them surpasses the other in some subsets; one of the subsets where \texttt{tb-FG19} takes the lead is Ma18, where that parser was developed and reported its results. This leads to the question whether developing and evaluating on a given subset of treebanks could induce bias in favor of those treebanks.
While the available data is not enough to give an answer in this specific instance, we can draw similar conclusions either way. If this were the case, it would mean that in the context of multilingual, language-agnostic parsers, and when data for a wide range of languages is available, it would be advisable to go beyond separating development and test sets for each language or treebank, and instead use different languages for development than for evaluation to avoid this kind of bias. Conversely, if this were not the case, it would mean that we could choose one of the human-defined subsets and obtain state-of-the-art results for one parser or the other, purely by chance. This makes us reflect about using a single subset of treebanks to justify the superior performance of a model, and might again make advisable to develop and test on different subsets - to reduce the element of chance.

\paragraph{Experimentation with data scarcity} 

For the D21 subset,
centered exclusively on extremely low-resource treebanks, the error reduction computed between the best performing parser (\texttt{gb-DM17}) and the control parser (\texttt{sl-S+19}) is the lowest among all tested treebanks. As mentioned above, the opposite happens for the easiest subsets. Yet, we feel these type of subsets would not be optimal either for evaluating parsers in a general sense, as they might not capture how a given parser can fully exploit its learning capabilities. Overall, the evaluation on this setup seems more volatile. We see a few differences between the predicted and gold PoS tags setups, causing even changes in the parsing ranking. For instance, \texttt{sl-S+19} outperforms \texttt{tb-FG19} in the gold setup by a clear margin, an issue that does not arise in any other subset.

\section{Experiment 2: \emph{many} parsers, \emph{many} randomized subsets}\label{section-experiment-2}

This experiment can be seen as a re-definition of Experiment 1 at a large scale. Above, we compared a few competitive parsers only on a handful of subsets of treebanks that were human-defined, and observed different trends. Yet, this is a limited view of the problem. If we take as reference the CoNLLU 2018 Shared Task \cite{zeman-etal-2018-conll} and the 82 treebanks that were evaluated, considering subsets of size 10 (meaning that each is composed of 10 different treebanks), we would obtain up to $\sim 2.13\times 10^{12}$ possible combinations. Many of those subsets will not be a representative sample of languages, but we already saw
that there are subsets that are used in parsing as a benchmark that are not either, and that even when they are considered representative, the criteria varies, and the parsing performance, differences among parsers and error reductions vary too. Here, we generate subsets, similar in size to typical human-defined ones, and see how subset differences affect
parsing rankings.

\begin{table}[phbt!]
\centering
    \scriptsize
    \addtolength{\tabcolsep}{-1.7pt}
    \begin{tabular}{l|ccccc}
        \textbf{Parser}&
        \textbf{Best rank} &
        \textbf{Worst rank} &
        $\boldsymbol{\mu}$ &
        $\boldsymbol{\tilde{\mu}}$ &
        $\boldsymbol{\sigma}$ \\
        \hline
        HIT-SCIR & 1 & 6 & 1.14 & 1.00 & 0.54 \\ 
        UDPipe Future & 1 & 12 & 4.38 & 4.00 & 1.65 \\ 
        TurkuNLP & 1 & 12 & 3.97 & 4.00 & 1.53 \\
        LATTICE & 1 & 13 & 4.99 & 5.00 & 2.27 \\ 
        ICS PAS & 2 & 13 & 4.71 & 5.00 & 2.01 \\ 
        CEA LIST & 1 & 13 & 6.12 & 6.00 & 2.28 \\ 
        Stanford & 1 & 21 & 6.30 & 6.00 & 3.47 \\ 
        Uppsala & 1 & 13 & 6.62 & 7.00 & 2.36 \\  
        NLP-Cube & 2 & 20 & 10.02 & 10.00 & 1.79 \\  
        AntNLP & 3 & 18 & 9.94 & 10.00 & 1.82 \\  
        ParisNLP & 4 & 20 & 10.39 & 11.00 & 1.62 \\  
        SLT-Interactions & 2 & 23 & 11.28 & 11.00 & 3.91 \\  
        IBM NY & 2 & 20 & 13.05 & 13.00 & 1.64 \\  
        LeisureX & 8 & 20 & 14.36 & 14.00 & 1.81 \\  
        UniMelb & 8 & 19 & 13.80 & 14.00 & 1.13 \\  
        KParse & 9 & 22 & 16.78 & 17.00 & 1.35 \\  
        Fudan & 10 & 22 & 17.16 & 17.00 & 1.60 \\  
        BASELINE UDPipe & 13 & 22 & 18.08 & 18.00 & 1.11 \\  
        Phoenix & 13 & 22 & 18.69 & 19.00 & 1.07 \\  
        CUNI x-ling & 2 & 22 & 19.24 & 20.00 & 2.21 \\  
        BOUN & 16 & 23 & 20.81 & 21.00 & 0.79 \\  
        ONLP lab & 20 & 25 & 22.57 & 23.00 & 0.62 \\  
        iParse & 9 & 25 & 22.36 & 23.00 & 2.76 \\  
        HUJI & 21 & 25 & 23.63 & 24.00 & 0.89 \\  
        ArmParser & 22 & 25 & 24.62 & 25.00 & 0.59 \\  
        SParse & 26 & 26 & 26.00 & 26.00 & 0.00 \\  
    \end{tabular}
    \caption{Ranking stats for LAS and the parsers of the \citet{zeman-etal-2018-conll} Shared task, over the 1 million random subsets. Table sorted by $\tilde{\mu}$ (the median).}
    \label{tab:ranking-metricas}
    \vspace*{0.25cm} 

        \centering
        \includegraphics[width=\columnwidth]{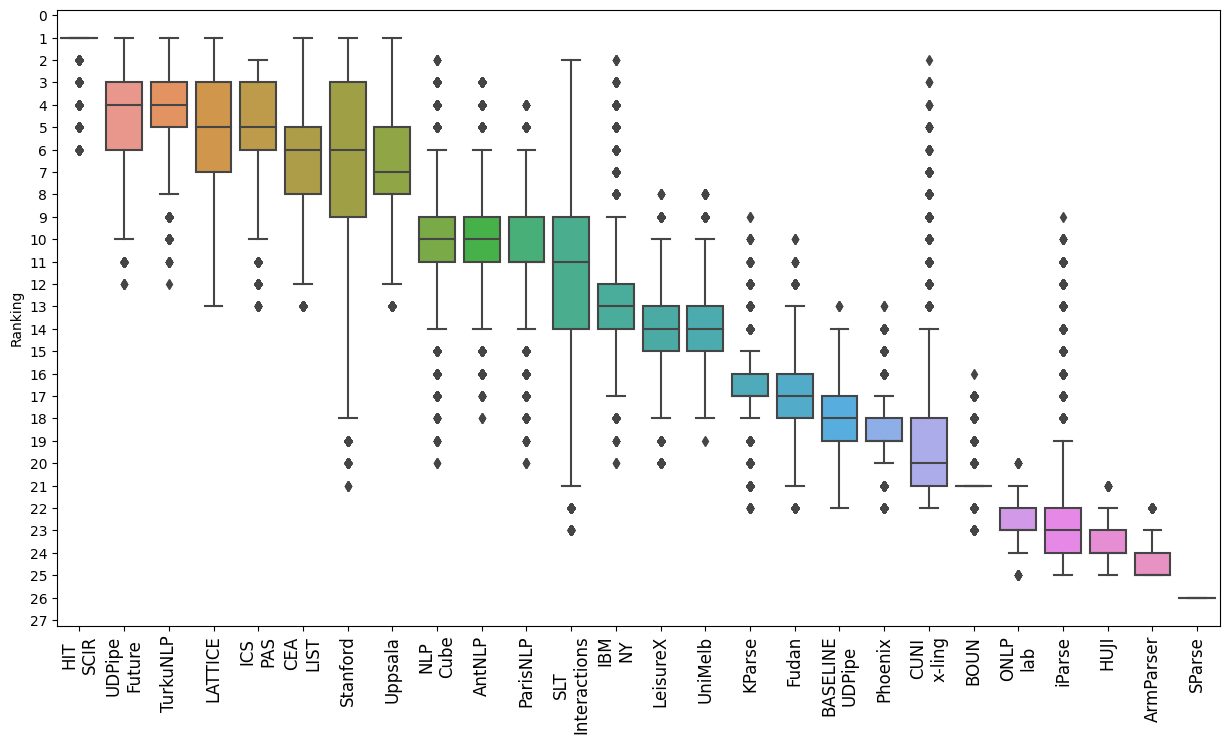}     \\
        \captionof{figure}{Corresponding box plot for Table \ref{tab:ranking-metricas}. For an easy correspondence with the table, the x-axis (from left to right) is sorted as the Table is (i.e. by $\tilde{\mu}$).}
        \label{fig:-boxplot-random-subsets}

\end{table}

\begin{table}[phbt!]
\centering
    \scriptsize
    \addtolength{\tabcolsep}{-1.7pt}
    \begin{tabular}{l|ccccc}
        \textbf{Parser}&
        \textbf{Best rank} &
        \textbf{Worst rank} &
        $\boldsymbol{\mu}$ &
        $\boldsymbol{\tilde{\mu}}$ &
        $\boldsymbol{\sigma}$ \\
        \hline
        TurkuNLP & 1 & 7 & 1.51 & 1.00 & 0.78 \\ 
        HIT-SCIR & 1 & 6 & 2.37 & 2.00 & 1.13 \\ 
        ICS PAS & 1 & 13 & 3.83 & 4.00 & 1.37 \\
        UDPipe Future & 1 & 8 & 3.62 & 4.00 & 0.98 \\ 
        Stanford & 1 & 15 & 4.21 & 4.00 & 1.75 \\ 
        LATTICE & 1 & 11 & 6.23 & 6.00 & 1.02 \\ 
        CEA LIST & 2 & 12 & 6.58 & 6.00 & 0.99 \\ 
        ParisNLP & 5 & 16 & 8.87 & 9.00 & 0.91 \\  
        AntNLP & 3 & 14 & 8.59 & 9.00 & 0.95 \\  
        SLT-Interactions & 2 & 20 & 10.09 & 10.00 & 2.25 \\  
        LeisureX & 7 & 18 & 11.54 & 11.00 & 1.21 \\  
        UniMelb & 7 & 16 & 11.27 & 11.00 & 0.82 \\  
        BASELINE UDPipe & 10 & 20 & 14.53 & 14.00 & 1.09 \\  
        NLP-Cube & 6 & 22 & 15.07 & 14.00 & 2.81 \\  
        Phoenix & 10 & 20 & 14.89 & 15.00 & 1.15 \\  
        KParse & 9 & 21 & 15.54 & 16.00 & 1.95 \\  
        CUNI x-ling & 4 & 21 & 17.43 & 18.00 & 1.60 \\  
        BOUN & 12 & 22 & 18.22 & 18.00 & 1.33 \\  
        Fudan & 9 & 22 & 17.34 & 18.00 & 1.94 \\  
        iParse & 9 & 25 & 19.12 & 20.00 & 3.19 \\  
        HUJI & 15 & 25 & 20.45 & 21.00 & 1.02 \\  
        ArmParser & 19 & 25 & 22.10 & 22.00 & 0.95 \\  
        Uppsala & 19 & 25 & 23.29 & 23.00 & 0.66 \\  
        IBM NY & 16 & 25 & 23.44 & 24.00 & 0.82 \\  
        ONLP lab & 22 & 25 & 24.88 & 25.00 & 0.37 \\  
        SParse & 26 & 26 & 26.00 & 26.00 & 0.00 \\  
    \end{tabular}
    \caption{Ranking statistics for BLEX and the parsers of the \citet{zeman-etal-2018-conll} Shared task, over 1 million randomly generated subsets. Table sorted by $\tilde{\mu}$.}
    \label{tab:ranking-metricas-blex}
    \vspace*{0.25cm} 

        \centering
        \includegraphics[width=\columnwidth]{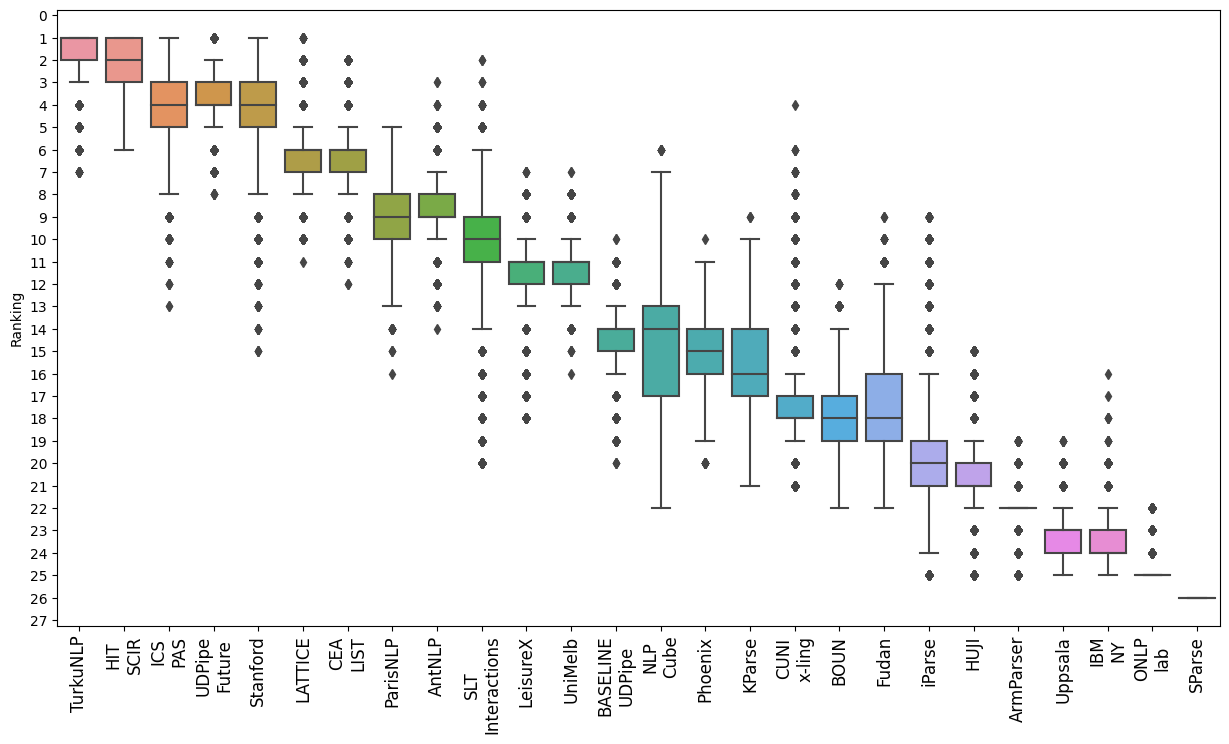}     \\
        \captionof{figure}{Corresponding box plot for Table \ref{tab:ranking-metricas-blex}. For an easy correspondence with the table, the x-axis (from left to right) is sorted as the table is (i.e. by $\tilde{\mu}$). }
        \label{fig:-boxplot-random-subsets-blex}

\end{table}

\begin{table*}[thbp!]
\centering
\addtolength{\tabcolsep}{-1.5pt}
\scriptsize
 \begin{tabular}{l|c|rc|rc}
 \multirow{2}{*}{\textbf{Parser}} & \multirow{2}{*}{\textbf{Avg. rank}}  & \multicolumn{2}{c|}{\textbf{Best subset outliers}} & \multicolumn{2}{c}{\textbf{Worst subset outliers}} \\ 
 &
 &
 \textbf{Rank} &
 \textbf{Treebanks} &
 \textbf{Rank} &
 \textbf{Treebanks} \\
 \hline
 UDPipe Future \cite{straka-2018-udpipe} & 4.38\textsubscript{$\pm$1.65} & 1 & \makecell{pl\textsubscript{sz}, ga\textsubscript{idt}, fro\textsubscript{srcmf}, sr\textsubscript{set}, kmr\textsubscript{mg},\\el\textsubscript{gdt}, en\textsubscript{ewt}, fr\textsubscript{gsd}, th\textsubscript{pud}, hy\textsubscript{armtdp}} & 12 & \makecell{bxr\textsubscript{bdt}, la\textsubscript{ittb}, hsb\textsubscript{ufal}, pcm\textsubscript{nsc}, th\textsubscript{pud},\\ro\textsubscript{rrt}, br\textsubscript{keb}, no\textsubscript{bokmaal}, sl\textsubscript{ssj}, sv\textsubscript{pud}} \\ 
 \hline
 Stanford \cite{qi-etal-2018-universal} & 6.30\textsubscript{$\pm$3.47} & 1 & \makecell{de\textsubscript{gsd}, cs\textsubscript{cac}, el\textsubscript{gdt}, he\textsubscript{htb}, es\textsubscript{ancora},\\sv\textsubscript{lines}, ja\textsubscript{modern}, bg\textsubscript{btb}, no\textsubscript{bokmaal}, cs\textsubscript{pud}} & 21 & \makecell{hsb\textsubscript{ufal}, pt\textsubscript{bosque}, ga\textsubscript{idt}, vi\textsubscript{vtb}, ko\textsubscript{kaist},\\sv\textsubscript{talbanken}, gl\textsubscript{treegal}, sme\textsubscript{giella}, hy\textsubscript{armtdp}, th\textsubscript{pud}} \\ 
 \hline
 SLT-Interactions \cite{bhat-etal-2018-slt} & 11.28\textsubscript{$\pm$3.91} & 2 & \makecell{sk\textsubscript{snk}, hsb\textsubscript{ufal}, uk\textsubscript{iu}, cs\textsubscript{fictree}, sv\textsubscript{pud},\\ja\textsubscript{gsd}, pt\textsubscript{bosque}, hr\textsubscript{set}, fa\textsubscript{seraji}, sl\textsubscript{ssj}} & 23 & \makecell{hr\textsubscript{set}, fa\textsubscript{seraji}, pl\textsubscript{lfg}, kmr\textsubscript{mg}, de\textsubscript{gsd},\\hy\textsubscript{armtdp}, nl\textsubscript{alpino}, th\textsubscript{pud}, sme\textsubscript{giella}, fro\textsubscript{srcmf}} \\
\end{tabular}
\caption{Qualitative results of subsets that cause anomalous rankings of parsers in Experiment 2.}
\label{tab:exp2-concrete-subset-examples}
\end{table*}

Similarly to Experiment 1, we do not analyze here algorithms and parsing architectures, or their correctness, but the appropriateness of evaluation procedures. In this context, some shared-task systems have reported bugs in their pipeline: this was mostly evident for some systems that consistently ranked in the last positions (see Table  \ref{tab:ranking-metricas}), but not so noticeable for high-scoring ones, such as the Stanford system \cite{qi-etal-2018-universal}, which later reported a preprocessing bug that affected the low-resource treebanks more. Thus, multi-treebank evaluation procedures should also be robust for systems suffering bugs that affect treebanks differently.

\subsection{Experimental setup}

We compare the available results of the 26 parsers\footnote{The parsers are not necessarily a diverse sample of parsing models (many are based on \texttt{gb-DM17}) but they are a realistic sample of a ranking of parsers made in a real shared task. For model representativity, we refer the reader to Experiment 1.} that participated in the CoNLLU Shared Task  \cite{zeman-etal-2018-conll}, sampling random subsets over the 82 evaluated treebanks. We generate 1 million random subsets made of size 10,\footnote{The subset size is in the range of those of Experiment 1.} and  we do not control the subsets' content (e.g. subsets with higher presence of a language or family).

\subsection{Results}

The pair Table \ref{tab:ranking-metricas} - Figure \ref{fig:-boxplot-random-subsets} shows statistics about the 26 parsers that participated in the shared task and the 1 million randomly generated subsets. Some (top) parsers show a stable performance. For instance, the HIT-SCIR parser \cite{che-etal-2018-towards} mostly ranks at the first position, except for a few outliers that show that the parser potentially could go down as far as the 6th position. This tendency is also observed in a few other - and worse-performing - systems, such as Kparse \cite{kirnap-etal-2018-tree} or Phoenix \cite{wu-etal-2018-multilingual}. 

However, for many other parsers the variability is larger. The interquartile range of the UDpipe-Future system (2nd place) \cite{straka-2018-udpipe} is small (from 3rd to 6th), but its fourth quartile (excluding outliers) ranges between the 6th and the 10th position. The situation is almost identical for the next 4 averaged best performing systems. Across the board, there are even more severe examples, such as Stanford \cite{qi-etal-2018-universal} (7th place), whose interquartile range spans from the 3rd to the 9th position; or the SLT-Interactions parser \cite{bhat-etal-2018-slt} whose first quartile ranges from 2nd to 9th, while its fourth quartile ranges from 14th to 21th. Exemplifying it with the Stanford system, we will also discuss below how randomized multi-treebank evaluation would have been useful to detect the anomalous performance on low-resource treebanks, that later on turned out to be a bug, or on the other hand how a weak subset selection could cause potential anomalous performances or bugs to go unnoticed.

\begin{table}[thbp!]
\centering
\scriptsize
 \begin{tabular}{p{1.5cm}|cccc}
 \textbf{Parser}&\textbf{\#Outliers}&\textbf{Avg. size}&\textbf{$\mathbfcal{R}$(lr)}&\textbf{$\mathbfcal{R}$(Slavic)}\\
 \hline
 UDPipe&10 best&\hfil168.87&\hfil0.30&\hfil0.17\\
 Future&10 worst&\hfil157.51&\hfil0.39&\hfil0.21\\
 \hline
 \multirow{2}{*}{Stanford}&10 best&\hfil221.03&\hfil0.19&\hfil0.23\\
 &10 worst&\hfil124.44&\hfil0.61&\hfil0.18\\
 \hline
 SLT-&10 best&\hfil203.92&\hfil0.30&\hfil0.36\\
 Interactions&10 worst&\hfil146.72&\hfil0.39&\hfil0.14\\
\end{tabular}
\caption{Quantitative study expanding Table \ref{tab:exp2-concrete-subset-examples}. $\mathcal{R}$ refers to the average ratio across subsets of the presence of low-resource (lr) and Slavic treebanks.}
\label{tab:quantitative-concrete-subset-examples}
\end{table}

\paragraph{Subsets that cause anomalous results}

Table \ref{tab:exp2-concrete-subset-examples} shows a few examples of parsers and subsets that caused atypical results. For each parser, we show an advantageous and a disadvantageous subset, randomly picked among those for which a parser obtained its best and worst rankings.

A qualitative analysis of these results yields several insights. For the first two parsers, the advantageous and disadvantageous outliers are linguistically diverse, but there is a clear trend that the disadvantageous subsets are heavily biased towards small treebanks: for both of the parsers, the favorable subset contains only 2 treebanks that are low-resource according to the shared task criteria,
while the disfavorable one contains 6 and 5, respectively. This is very unlikely to happen by chance: the probability of randomly drawing a subset with 6 or more low-resource treebanks is 0.00214, and with 5 or more, 0.01538 (this is calculated from a hypergeometric distribution with parameters $N=82$, the total number of treebanks, $K=21$, the number of low-resource treebanks, and $n=10$, the number of treebanks per subset). Thus, this variability seems to owe to the fact that the UDPipe Future \cite{straka-2018-udpipe} and Stanford \cite{qi-etal-2018-universal} parsers struggle (relatively to competitors) when training data is scarce. The situation is different for the third parser considered. In this case, there are no substantial differences in treebank size (2 vs. 3 low-resource treebanks) but instead there is a clear linguistic pattern: the advantageous subset has a heavy bias towards Slavic languages (6 out of the 10 languages are Slavic, compared to 2 in the disadvantageous subset - and the probability of choosing a subset with 6 or more Slavic languages by chance is 0.00043, from a hypergeometric distribution with parameters $N=82, K=17, n=10$). This seems to reflect that the SLT-Interactions parser \cite{bhat-etal-2018-slt} is especially adequate for Slavic languages. It is worth noting that the authors did not implement any language-specific adaptation or report anything in the paper that suggests that they specifically addressed these languages, so this serves as an example that a parser can show linguistic biases towards certain language families even if it has been developed in a language-agnostic way.

We propose a complementary quantitative analysis in Table \ref{tab:quantitative-concrete-subset-examples}. We randomly take 10 of the best and worst performing subsets for the above studied parsers and compute, across subsets, the average size of treebanks, the presence of low-resource treebanks, and the presence of Slavic languages. The analysis confirms the bias towards rich-resource treebanks for the Stanford parser, and towards Slavic languages for SLT-Interactions (while not being biased towards rich- or low-resouce treebanks). On the other hand, the hypothesis of UDpipe Future being biased towards rich-resource languages is not clearly confirmed by this analysis.

To sum up, this reinforces the idea (hypothesized in papers like \citet{deLhoneux2016UDTS}) that both treebank sizes and linguistic factors are important for a treebank subset to be representative; and highlight that the latter can have a huge influence even in parsers that have been developed without specific language families in mind.

\paragraph{More robust metrics?} LAS and UAS are the most popular metrics to report dependency parsing performance.
Yet, there are other metrics, such as CLAS\footnote{CLAS: It ignores selected relations which attach function words to content words.}, MLAS\footnote{MLAS: It is inspired by the CLAS metric, and extended with evaluation of POS tags and morphological features.} or BLEX\footnote{BLEX (bi-lexical dependency score): it combines content-word relations with lemmatization.}, but they have not been widely adopted (maybe because they have a not so straightforward interpretation). Yet, from Experiment 2 we observed that some of these metrics, especially BLEX, produced narrower standard deviations and more stable rankings. 
We leave interpretations of this phenomenon as an open question for future work, but refer the reader to (Table \ref{tab:ranking-metricas-blex}, Figure \ref{fig:-boxplot-random-subsets-blex}) and (Table \ref{tab:ranking-metricas}, Figure \ref{fig:-boxplot-random-subsets}),
which show a summary of the ranking statistics for the BLEX and LAS metrics, respectively, on the 26 parsers that participated in the ConLLU Shared task 2018 \cite{zeman-etal-2018-conll}. Overall, but especially for the top parsers, BLEX results produce more stable rankings and narrower interquartile ranges.

\section{Discussion}

We have designed two experiments that revealed issues of relying on a single subset of treebanks for parsing evaluation. More particularly, we have shown that: (i) existing human-defined subsets show high variability in terms of rankings and performance across parsers, (ii) parsers that have been developed on a concrete subset might be biased towards performing better on that subset, (iii) it is relatively easy to come up with subsets that generate different parsing rankings, (iv) this can even happen across subsets that have been purposefully defined to be representative, (v) both linguistic typology and resource size have a large influence in the variability of results between parsers, and (vi) linguistic factors can be crucial even when parsers are designed in a language-agnostic way.

Overall, some advice can be given: (a) claims that ``parser X is more accurate than parser Y'' can be weak even on carefully selected samples of UD treebanks (and perhaps it is recommendable to consider metrics that take into account dimensions such as speed and efficiency), (b) for language-agnostic parsers, it is worth noting that there can still be biases towards certain linguistic families, and (c) for such parsers, it can be advisable to develop on one set of treebanks and evaluate on another, to avoid bias in favor of the languages used for development.

Finally, there are aspects that we did not study in this piece of work, but that could affect the robustness of parsing evaluation as well, e.g., automatically \emph{versus} manually annotated treebanks, and interactions between language and treebank properties (e.g. morphological complexity, dependency distance, \dots) and parsing models.

\section{Conclusion}

Different subsets of treebanks have been proposed to try to capture the essence of the whole set of UD treebanks, so that
the performance of parsers in such subsets would be representative of that obtained in the full set. We have empirically shown limitations of this approach, and also how establishing guidelines for good treebank selection can be hard, although some bad practices can be avoided.

\section*{Acknowledgments}

This work was supported by a 2020 Leonardo Grant for Researchers and Cultural Creators from the FBBVA,\footnote{FBBVA accepts no responsibility for the opinions, statements and contents included in the project and/or the results thereof, which are entirely the responsibility of the authors.} as well as by the European Research Council (ERC), under the European Union’s Horizon 2020 research and innovation programme (FASTPARSE, grant agreement No 714150). The work is also supported by ERDF/MICINN-AEI (SCANNER-UDC, PID2020-113230RB-C21), by Xunta de Galicia (ED431C 2020/11), and by Centro de Investigación de Galicia ‘‘CITIC’’ which is funded by Xunta de Galicia, Spain and the European Union (ERDF - Galicia 2014–2020 Program), by grant ED431G 2019/01.


\bibliography{anthology,custom}

\clearpage
\appendix

\section{Experiment 1: models, resources, and hyperparameters}\label{appendix-hyperparameters}

\noindent To train the models, we used 2 NVIDIA GeForce RTX 2080 Ti@11GB and an Intel® Core™ i7-9700K@3.60GHz×8. Training times usually took from 1 to 7 hours, depending on the parsing model and the treebank training size. The three used parsers and the UD treebanks have free software licenses that allow free use and distribution.

\noindent Tables \ref{tab:hyperparameters-supar}, \ref{tab:hyperparameters-syntacticPointer} and \ref{tab:hyperparameters-dep2label} show the hyperparameters used for the \texttt{gb-DM17} \cite{dozat-etal-2017-stanfords} (using the \texttt{supar} software package), \texttt{tb-FG19} \cite{fernandez-gonzalez-gomez-rodriguez-2019-left} (using the \texttt{syntacticpointer} package) and \texttt{sl-S+19} parsers \cite{strzyz-etal-2019-viable} (using the \texttt{dep2labels} package), respectively.

\begin{table}[hpbt]
\centering
\small
\tabcolsep=0.08cm
\begin{tabular}{lclc}
\hline
Hyperparameter & Value & Hyperparameter & Value \\
 \hline
n\_char\_hidden & 100 & $\nu$ & $.9$ \\
n\_feat\_embed & 100 & $\epsilon$ & $1^{-12}$ \\
embed\_dropout & $.33$ & weight\_decay & 0 \\
n\_lstm\_hidden & 400 & clip & 5.0 \\
n\_lstm\_layers & 3 & min\_freq & 2 \\
encoder\_dropout & $.33$ & fix\_len & 20 \\
n\_arc\_mlp & 500 & decay & $.75$ \\
n\_rel\_mlp & 100 & decay\_steps& 5000 \\
mlp\_dropout & $.33$ & update\_steps & 1 \\
encoder & lstm & feats & ['tag', 'char'] \\
\hline
\end{tabular}
\caption{Hyperparameters used to train the \texttt{supar} models. In the case of Ancient Greek the hyperparameter \textit{n\_embed} is 100.}
\label{tab:hyperparameters-supar}
\end{table}

\begin{table}[hpbt]
\centering
\small
\tabcolsep=0.06cm
\begin{tabular}{lclc}
\hline
 Hyperparameter & Value & Hyperparameter & Value \\
 \hline
 model & L2RPtr & --learning\_rate & 0.001 \\
 word\_dim & 300 & --lr\_decay & 0.999997 \\
 char\_dim & 100 & --beta1 & 0.9 \\
 pos & true & --beta2 & 0.9 \\
 rnn\_mode & FastLSTM & --grad\_clip & 5.0 \\
 encoder\_layers & 3 & --loss\_type & token \\
 decoder\_layers & 1 & --warmup\_steps & 40 \\
 hidden\_size & 512 & --reset & 20 \\
 arc\_space & 512 & --weight\_decay & 0.0 \\
 type\_space & 128 & --unk\_replace & 0.5 \\
 p\_in & 0.33 & --beam & 5 \\
 p\_out & 0.33 & --char\_embedding & random \\
 p\_rnn & [0.33, 0.33] & --opt & adam \\
 prior\_order & inside\_out & --batch\_size & 32 \\
 grandPar & false & --num\_epochs & 600 \\
 sibling & false & & \\
 activation & elu & & \\
\hline
\end{tabular}
\caption{Hyperparameters used to train the \texttt{syntacticpointer} models.  Parameters specified from the configuration file on the left, and from the command line on the right.}
\label{tab:hyperparameters-syntacticPointer}
\end{table}

\begin{table}[phtb]
\centering
\small
\tabcolsep=0.06cm
\begin{tabular}{lcc}
\hline
 Hyperparameter & Value \\
 \hline
 cnn\_layer & 4 \\
 char\_hidden\_dim & 100 \\
 hidden\_dim & 800 \\
 dropout & 0.5 \\
 lstm\_layer & 3 \\
 bilstm & True \\
 learning\_rate & 0.02 \\
 lr\_decay & 0.05 \\
 momentum & 0.9 \\
 l2 & 0 \\
 gpu & True \\
\hline
\end{tabular}
\caption{Hyperparameters used to train the \texttt{dep2labels} models.}
\label{tab:hyperparameters-dep2label}
\end{table}

\section{Treebanks in each subset}\label{appendix-treebanks-per-subset}

In \S \ref{section-datasets-exp1}, we reviewed the related work and briefly discussed several human-defined subsets that were proposed in the past, according to a number of criteria, and that we used to report the results from our Experiment 1. Due to space reasons, we detail here in this appendix (Table \ref{tab:treebanks-per-set}) the specific treebanks that are part of each subset, and their sizes, for a better understanding of the particularities of each of them.

\begin{table*}[hbtp!]
\centering
\addtolength{\tabcolsep}{-1pt}
\small
\begin{tabular}{c c c c c c c c c c}
 & Size & Ma18 & Lh16 & AG20 & D21 & SA17 & Sm18 & Ku19 & Easy \\ \rowcolor[gray]{.95}
\hline
Ancient Greek (PROIEL) & 213K & & \checkmark & & & & \checkmark & & \\
Ancient Greek (Perseus) & 202K & & & \checkmark & & & \\ \rowcolor[gray]{.95}
Arabic (PADT) & 282k & & & & & & \checkmark & \checkmark & \\
Basque (BDT) & 121K & & & & & & & \checkmark & \\ \rowcolor[gray]{.95}
Belarusian (HSE) & 305K & & & & \checkmark & & & & \\ 
Bulgarian (BTB) & 156K & \checkmark & & & & & & & \checkmark \\ \rowcolor[gray]{.95}
Catalan (AnCora) & 546K & \checkmark & & & & & & & \checkmark \\ 
Chinese (GSD) & 123K & & \checkmark & \checkmark & & & \checkmark & \checkmark & \\ \rowcolor[gray]{.95}
Coptic (Scriptorium) & 48K & & & & & \checkmark & & & \\ 
Czetch (FicTree) & 167K & & & & & & & & \checkmark \\ \rowcolor[gray]{.95}
Czetch (PDT) & 1509K & \checkmark & \checkmark & & & & & & \\ 
Dutch (Alpino) & 208K & \checkmark & & & & \checkmark & & & \\ \rowcolor[gray]{.95}
English (EWT) & 254K & \checkmark & \checkmark & \checkmark & & & \checkmark & \checkmark & \\ 
Finnish (TDT) & 202K & & \checkmark & \checkmark & & & \checkmark & \checkmark & \\ \rowcolor[gray]{.95}
French (GSD) & 400K & \checkmark & & & & & & & \\ 
Galician (TreeGal) & 25K & & & & \checkmark & & & & \\ \rowcolor[gray]{.95}
German (GSD) & 292K & \checkmark & & & & & & & \\ 
Hebrew (HTB) & 161K & & \checkmark & \checkmark & & \checkmark & \checkmark & \checkmark & \\ \rowcolor[gray]{.95}
Hindi (HDTB) & 351K & & & & & & & \checkmark & \checkmark \\ 
Indonesian (GSD) & 120K & & & & & \checkmark & & & \\ \rowcolor[gray]{.95}
Italian (ISDT) & 298K & \checkmark & & & & \checkmark & & \checkmark & \checkmark \\ 
Japanese (GSD) & 193K & & & & & & & \checkmark & \\ \rowcolor[gray]{.95}
Kazakh (KTB) & 10K & & \checkmark & & & & & & \\ 
Korean (GSD) & 80K & & & & & & & \checkmark & \\ \rowcolor[gray]{.95}
Korean (Kaist) & 350K & & & & & & \checkmark & & \\ 
Lithuanian (HSE) & 5K & & & & \checkmark & & & & \\ \rowcolor[gray]{.95}
Marathi (UFAL) & 3K & & & & \checkmark & & & & \\ 
Norwegian (Bokmaal) & 310K & \checkmark & & & & \checkmark & & & \checkmark \\ \rowcolor[gray]{.95}
Old Church Slavonic (PROIEL) & 57K & & & & & \checkmark & & & \\ 
Old East Slavic (RNC) & 30K & & & & \checkmark & & & & \\ \rowcolor[gray]{.95}
Polish (LFG) & 130K & & & & & & & & \checkmark \\ 
Polish (PDB) & 350K & & & & & \checkmark & & & \\ \rowcolor[gray]{.95}
Romanian (RRT) & 218K & \checkmark & & & & & & & \\ 
Russian (GSD) & 98K & & & \checkmark & & & & & \\ \rowcolor[gray]{.95}
Russian (SynTagRus) & 1107K & \checkmark & & & & & \checkmark & \checkmark & \checkmark \\ 
Sanskrit (Vedic) & 27K & & & & & \checkmark & & & \\ \rowcolor[gray]{.95}
Slovenian (SSJ) & 140K & & & & & & & & \checkmark \\ 
Spanish (AnCora) & 560K & \checkmark & & & & & & & \checkmark \\ \rowcolor[gray]{.95}
Swedish (Talbanken) & 96K & & & & & & \checkmark & \checkmark & \\ 
Tamil (TTB) & 9K & & \checkmark & \checkmark & \checkmark & & & & \\ \rowcolor[gray]{.95}
Turkish (IMST) & 57K & & & & & & & \checkmark & \\ 
Uyghur (UDT) & 40K & & & \checkmark & & & & & \\ \rowcolor[gray]{.95}
Welsh (CCG) & 36K & & & & \checkmark & & & & \\ 
Wolof (WTB) & 44K & & & \checkmark & & & & & \\
\end{tabular}
\caption{Treebanks per set}
\label{tab:treebanks-per-set}
\end{table*}

\end{document}